\pdfoutput=1

\documentclass[10pt, a4paper]{article}

\usepackage{lrec-coling2024}
\usepackage{hyperref}
\usepackage{times}
\usepackage{latexsym}
\usepackage{multirow}
\usepackage{booktabs}
\usepackage{microtype}
\usepackage{multirow}
\usepackage{latexsym}
\usepackage{booktabs}
\usepackage{courier}
\usepackage{amsmath}
\usepackage{booktabs} 
\usepackage{siunitx} 
\usepackage{subfig}
\usepackage{placeins}
\usepackage{tikz}
\usepackage{hyperref}

\usepackage[T1]{fontenc}

\usepackage[utf8]{inputenc}

\usepackage{microtype}

\usepackage{inconsolata}

\title{Designing NLP Systems That Adapt to Diverse Worldviews}

\name{Claudiu Creangă$^{2,3,}$, Liviu P. Dinu$^{1,3}$} 

\address{  $^1$ Faculty of Mathematics and Computer Science  \\
  $^2$ Interdisciplinary School of Doctoral Studies, 
  $^3$ HLT Research Center \\
  University of Bucharest, Romania\\
         claudiu.creanga@s.unibuc.ro, ldinu@fmi.unibuc.ro}

\abstract{
Natural Language Inference (NLI) is foundational for evaluating language understanding in AI. However, progress has plateaued, with models failing on ambiguous examples and exhibiting poor generalization. We argue that this stems from disregarding the subjective nature of meaning, which is intrinsically tied to an individual's \textit{weltanschauung} (which roughly translates to worldview).  Existing NLP datasets often obscure this by aggregating labels or filtering out disagreement. We propose a perspectivist approach: building datasets that capture annotator demographics, values, and justifications for their labels. Such datasets would explicitly model diverse worldviews. Our initial experiments with a subset of the SBIC dataset demonstrate that even limited annotator metadata can improve model performance.
 \\ \newline }
\begin{document}

\maketitleabstract

\section{Introduction}

Natural Language Inference (NLI) lies at the heart of developing and evaluating language understanding in AI models. As Montague stated, entailment is "the basic aim of semantics" (Montague, 1970) and a lot of focus has been put in building models that score high on NLI datasets. Recently, two big issues emerged. Firstly, the research has reached a plateau on these datasets where the models perform almost as well as humans on samples where there is high human agreement, but perform poorly on samples with high entropy (from 0.9 to 0.5 accuracy) \cite{nie}. Secondly, models have poor generalisation abilities and their high score on in-distribution samples doesn't translate to a high score on out-of-distribution samples \cite{Bras}, \cite{zhou-bansal-2020-towards}, a sign that they actually use shallow heuristics rather than understanding. We will argue that these models lack a worldview and that using perspectivist approaches we can improve our solutions to both problems. A significant obstacle in this research is the scarcity of datasets that preserve annotator demographics and reveal their socio-political values, which are crucial for understanding their worldviews. Our code is open source and available to use on \href{https://github.com/anonymous}{Anonymous GitHub}. Our present work focuses on worldviews which are shaped culture, values, beliefs and less by demographic data. Several studies \cite{Orlikowski} have shown that demographic data alone is not a good predictor of annotator's views. 

\section{Related Work}

\citet{Basile} offer a valuable synthesis of prior research in perspectivist machine learning, clearly delineating two distinct approaches within the field: weak and strong perspectivism. Historically, the illusion of ground truth was ingrained in every NLP dataset. At start, NLP datasets were built only with a single annotator per label. That label was taken as the truth, even if the language was ambiguous or the annotator made a mistake. Then, a step forward was made when weak perspectivist research acknowledged the potential for disagreement and errors. It therefore adopted a multi-annotator approach to capture diverse perspectives. However, where there was human disagreement, this was solved either by aggregation (majority voting) or by filtering (removing low agreement samples).  But, by removing the low agreement samples, we remove a big part of human speech which is by its virtue ambiguous and, by aggregation, we obscure the rich diversity of valid interpretations inherent in human communication. Strong perspectivist research embraces linguistic diversity, recognizing that even when aiming for a single target label, models benefit from exposure to non-aggregated data (manifesto\footnote{https://pdai.info/}). In the literature we've found three main ways to embrace variation: 
\begin{itemize}
    \item \textbf{Multilabel categorical classification:} where a single model predicts multiple labels for each sample \cite{ferracane-etal-2021-answer}, \cite{jiang-marneffe-2022-investigating} and others;
    \item \textbf{Soft label classification:} where a single model is trained on the distribution of labels for a given sample and the model predicts that distribution \cite{1908.07086}, \cite{Uma} and others. 
    \item \textbf{Radical perspectivist:} where we train a model for each annotator so that a model learns the behaviour of one particular annotator \cite{Akhtar}. This requires the identification of each annotator, which few datasets have. 
\end{itemize}

Our research continues in the line of the radical perspectivist approach and takes it one step further, by including not only the demographics of the annotators, but also the values and beliefs which make the annotator's worldview.

\section{The goal of an NLP system}

NLP is a large field and uses of NLP systems can vary from translations to information extraction, summarization and others. But at its core, we can say that an NLP system aims to enable computers to understand and generate human language in a meaningful way. We argue that meaning is subjective and cannot be separated from a worldview. Our language influences how we perceive and understand the world and the other way around. The subjective nature of meaning doesn't entail that each annotator position is equally valid under the current paradigm (an extreme relativistic view that can have negative implications in some tasks like hate speech detection \cite{2403.02268}), but that each position is part of a worldview that should be dealt with, not erased.

\subsection{Meaning and Worldview}

Quine's radical indeterminacy \cite{Quine} shows that no sentence has only one meaning and there is no way to determine the one correct translation of a sentence in another language. This happens because meaning is embedded within the speaker's entire web of beliefs, culture, and how they experience the world. The term that better describes this concept is \textbf{Weltanschauung}, roughly translated by worldview. The main problem of NLP systems today is that it doesn't take into account the different worldviews in which a sentence can be interpreted. We cannot know the right label for large category of utterances if we don't contextualise it in a worldview. If we accept Quine's position that there's no such thing as \textbf{purely mental meanings} (mental dictionaries from which we take the definition of every word we use) and that what words mean is inextricably linked to how speakers behave and look at the world (\textit{weltanschauung}), then we can agree that the legacy NLP datasets do not offer the information we need. By \textbf{legacy datasets} we mean datasets that only have aggregated labels. But non-aggregated labels are not enough and we should go further and include those that do not offer labels by annotator id, demographic metadata and worldview metadata about each annotator. 

Here we need to recognize that while many inferences are influenced by worldview, certain types – such as those grounded in formal logic or mathematical reasoning – hold regardless of the annotator (\textbf{deductive inferences}). If we were to use logic to establish what inference means, we would denote it by this formula:

\begin{quote}

$\forall w \in W \, : \, (P(w) \rightarrow H(w)) $

\end{quote}

Which means that for every world \textit{\textbf{w}}, if the premise \textit{\textbf{P}} is true in world \textit{\textbf{w}}, then the hypothesis \textit{\textbf{H}} is also true in world \textit{\textbf{w}}. Only deductive inferences can pass this type of rigour:
\begin{quote}




\textbf{Premise}: All men are mortal and Socrate is a man.

\textbf{Hypothesis}: Socrate is mortal. 

\textbf{Label}: Entailment. 
\end{quote}

Here, no matter what we mean by Socrate and men, the inference is still valid. Compared to deductive inferences, \textbf{inductive inferences} need to pass a lower bar. Traditionally the standard was what a "common man" would assume to be true or false about an utterance. The creators of datasets sometimes give annotators instructions on how to evaluate utterances \cite{bowman-etal-2015-large}. In case of SNLI, entailment meant a "definitely true description" and neutral "might be a true description". In case of MNLI the instructions for entailment was "definitely correct" and "might be correct" for neutral. In \citealt{Gubelmann} the threshold is much more lax, a "good reason" is sufficient for entailment and it is given the following example:

\begin{quote}
\textbf{Premise}: The streets are wet.

\textbf{Hypothesis}: It has rained. 

\textbf{Label}: Entailment. 
\end{quote}

Firstly, we argue that even a simple inference like this cannot be made without considering the annotator's context. For instance, an annotator living in a town where streets are washed every morning might attribute wet streets to cleaning rather than rain as his first thought. Secondly, providing annotators with instructions biases the results, as it attempts to override their individual understanding of entailment and contradiction. This prevents the study from authentically reflecting common language use which comes in different varieties. And NLP systems should reflect how humans naturally speak.

\subsection{Building a worldview-annotated dataset}

Consider the following basic pair of sentences:

\begin{quote}
\textbf{Premise}: It is dark outside.

\textbf{Hypothesis}: It is dangerous to go outside. 
\end{quote}

There is no right label for this pair and the annotator would use his life experiences to annotate it. A woman would be more likely to annotate it as an entailment. Many men who live in high crime areas would potentially do the same, while others would see no good reason to be dangerous outside if it is dark. Recently built perspectivist datasets, such as ChaosNLI \cite{ynie2020chaosnli} which collects 100 annotations for each label, but then proceeds to remove the worldviews of annotators and provide only a distribution (i.e. 30\% E, 50\% N, 20\% C) make the same mistake of weak perspectivist research. Relying solely on a distribution, one cannot learn in which worldview this statement is entailment or not and if we don't know the annotator backgrounds and how diverse they are, we risk to capture only a limited range of potential interpretations. 

\begin{figure*} 
  \centering
  \includegraphics[width=0.8\linewidth]{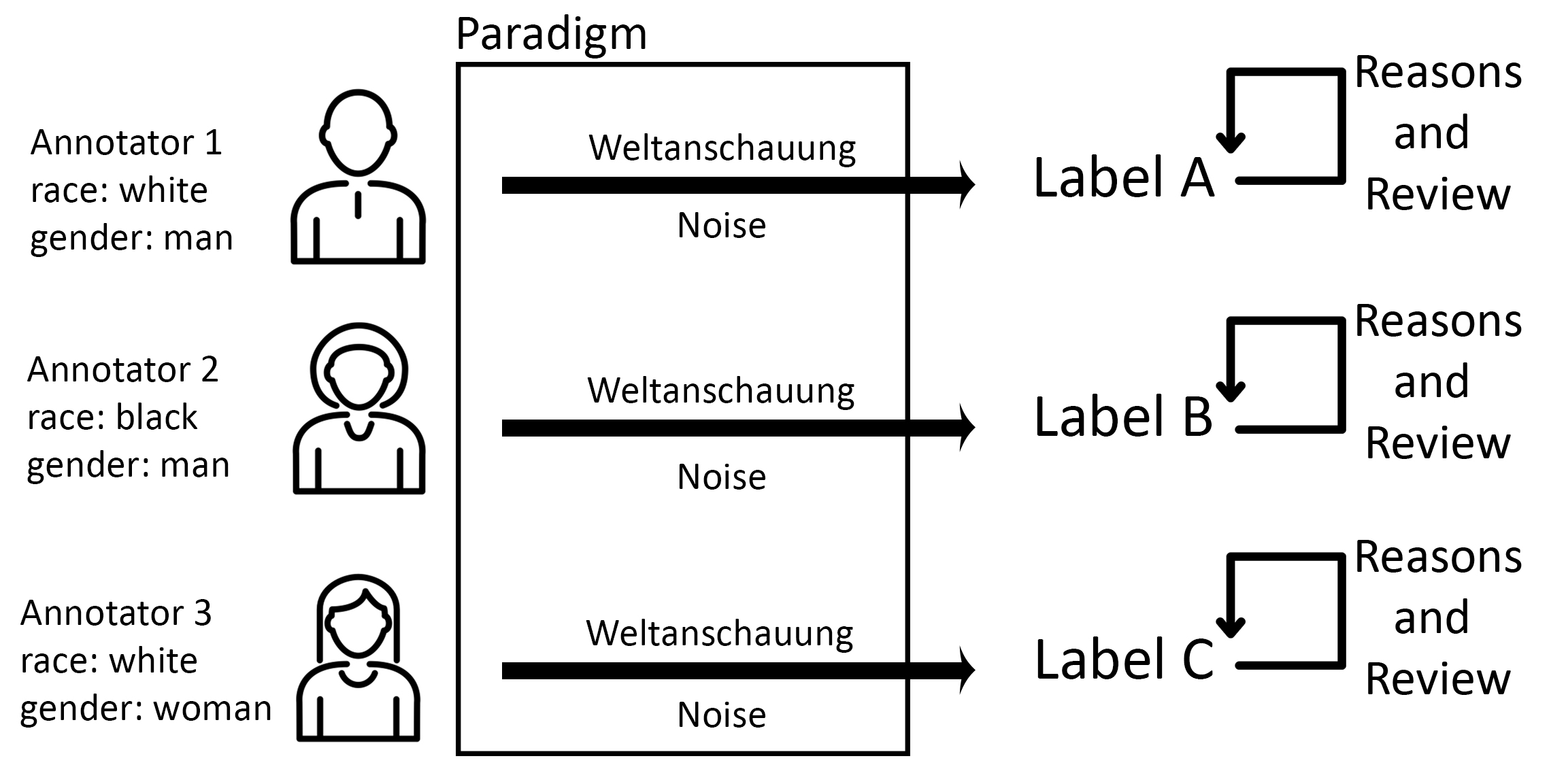}
  \caption{Building a worldview-annotated dataset. It is necessary to have a diverse pool of annotators aligned to the task. Metadata should be collected about each of these annotators: demographic and values. Each annotator should label items according to their worldview, while being mindful of potential "noise" –  errors caused by factors like inattention, which are unrelated to their perspective. To mitigate noise and preserve valid interpretations, annotators should provide justifications for their labels.  They will then self-review their labels based on their own explanation.}
  \label{fig:annotator}
\end{figure*}

In \autoref{fig:annotator} we explain how we can build worldview-annotated datasets. It is necessary to have a diverse pool of annotators aligned to the task. For instance, a BioNLP task would benefit from annotators with expertise in health sciences. Metadata should be collected about each of these annotators: demographics and values. Worldviews are part of a paradigm, a set of principles shared by every annotator at a certain point in time. Each annotator should label items according to their worldview, while being mindful of potential "noise" –  errors caused by factors like lack of attention, which are unrelated to their perspective. To mitigate noise and preserve valid interpretations, annotators should provide justifications for their labels. They will then self-review their labels based on their own explanation and reconsider the label or not. This self-review step should remove some noise. The \textbf{justifications} are important for a second reason. How can we address the scenario where two annotators assign the same label, but their justifications diverge so significantly that the apparent agreement is misleading? This case suggests that reasons must be kept in the dataset and if reasons diverge significantly there should be different categories in each label. A model that learns a worldview should arrive at a label through the same reasoning as the annotator. Unfortunately there is no public dataset where reasons are available, but we aim to build one and we hypothesize that it would help the models in learning a worldview. 

Training a model in this way means modelling a worldview in which the model doesn't pretend universality and is by essence local. We hypothesize that this grounding in a worldview will make the model more consistent in its judgements and generalise better. 

\section{Modelling a worldview}

Unfortunately, the demographic metadata of annotators is removed in most datasets and there are very few datasets that collected annotators' social and political values in order to create a worldview. However, this is changing, with at least four datasets that we are aware of now including more information: \citealt{Chulvi}, \citealt{sachdeva-etal-2022-measuring}, \citealt{sap} and \citealt{Hettiachchi}. In \cite{sap-etal-2022-annotators} we see that annotators with stronger racist beliefs demonstrate a  tendency to mislabel African American English (AAE) as toxic, while being less likely to identify anti-Black language as harmful, suggesting that even only demographic metadata can be helpful. To the best of our knowledge, SBIC \cite{sap} is the most complete dataset in regards to including both annotator demographics and the political beliefs of the annotator (liberal or conservative). Their annotators are from U.S. and Canada and, although they are gender and age-balanced, the ethnicity is not diverse, there are a lot more white (82\%) than any other group (4\% Asian, 4\% Hispanic, 4\% Black). They annotated posts from Reddit and Twitter for offensiveness, intent to offend, sexual references and, if it was offensive, which groups did it target. 

We used this dataset for our goal of building models with demographic and  political medatada. Because the dataset was built with detecting Social Bias Frames in mind, they had 263 different annotators and at most 3 annotators per post. For our use case we had to find annotators from different backgrounds that annotated the same posts. The biggest subset that we could find from this dataset was for 2 annotators, man (liberal, white) and woman (mod-conservative, white), which together annotated 290 samples (worker ids are hidden due privacy concerns). 

Given it is a small dataset, we are aware of the limitations of our results. We used K-Fold validation (10) to reduce the risk of overfitting and we used a pre-trained DeBERTaV3 model \cite{he2021deberta} followed by a fully connected layer for predictions. We used a learning rate of 5e-5 for 3 epochs where we trained only the output layer and then a small learning rate of 2e-5 for one epoch where we trained the output layer and the last layer from the pre-trained model. The target label is if the post is offensive or not. 

\begin{table}
  \caption{F1 scores of the three types of training: training on aggregated data, using 2 models for each annotator with and without metadata.}
  \label{table:results}
  \begin{tabular}{ccc}
    \toprule
     Training Type & Val score & Test score \\
    \midrule
    \texttt Aggregated data & 0.3 & 0.3  \\
    \texttt 2 models no metadata & 0.37 & 0.36  \\
    \texttt 2 models and metadata & 0.38 & 0.38  \\
    \bottomrule
  \end{tabular}
\end{table}

As shown in (\autoref{table:results}), leveraging annotator metadata provides a boost in performance. Training on aggregated data without metadata yields a test F1 score of 0.3.  While splitting the data by annotator, training 2 separated models and aggregating the outputs afterwards improves the F1 score to 0.36.  Instead, if we also concatenate the annotators' metadata to the input before the encoding layers, we increase the F1 score to 0.38. Even though the dataset is small, there is an improvement when we consider the annotators' background.

\section{Conclusion}

Traditional NLP approaches have exhibited limitations in both generalization and performance on challenging examples within NLI datasets. Part of these shortcomings stem from the omission of the subjective nature of meaning and the diverse worldviews that shape individual interpretations of language. For inductive inferences, the notion of a universally "correct" label detached from a worldview is misleading.

We propose that embracing perspectivist approaches and building worldview-annotated datasets is crucial for advancement in NLP. Such datasets must capture annotator justifications, demographic information, and the values that comprise their worldview, offering a richer understanding of linguistic variation. Such datasets are missing at the moment, but initial experiments with a subset of the SBIC data support our hypothesis: even with limited demographics and a basic political orientation label, incorporating annotator metadata improves model performance.

\section*{Acknowledgements}
This work was partially supported by a grant on Machine Reading Comprehension from Accenture Labs and by the POCIDIF project in Action 1.2. ``Romanian Hub for Artificial Intelligence''.

\bibliography{custom}
\bibliographystyle{acl_natbib}

\end{document}